\documentclass{llncs}

\usepackage{todonotes}
\usepackage[utf8]{inputenc}
\usepackage{url}
\usepackage{glossaries}
\glsdisablehyper
\newacronym{nlp}{NLP}{Natural Language Processing}
\newacronym[longplural={Recurrent Neural Networks}]{rnn}{RNN}{Recurrent Neural Network}
\newacronym{mlstm}{mLSTM}{Multiplicative Long Short-Term Memory}
\newacronym{gru}{GRU}{Gated Recurrent Unit}
\newacronym{lida}{LIDA}{Late Inter-Document Averaging}
\newacronym{cida}{CIDA}{Continual Inter-Document Averaging}
\newacronym{ida}{IDA}{Inter-Document Attention}
\newacronym{iida}{InIDA}{Inter- and Intra-Document Attention}
\newacronym{svm}{SVM}{Suport Vector Machine}

\usepackage[normalem]{ulem}

\title{Inter and Intra Document Attention for Depression Risk Assessment}
\author{Diego Maupomé, Marc Queudot, Marie-Jean Meurs}
\institute{Universit\'e du Qu\'ebec \`a Montr\'eal, Montr\'eal, QC, Canada}
\date{~}

\begin{document}

\maketitle

\begin{abstract}
We take interest in the early assessment of risk for depression in social media users.
We focus on the eRisk 2018 dataset, which represents users as a sequence of their written online contributions.
We implement four RNN-based systems to classify the users.
We explore several aggregations methods to combine predictions on individual posts.
Our best model reads through all writings of a user in parallel but uses an attention mechanism to prioritize the most important ones at each timestep.

\end{abstract}

\section{Introduction}

In 2015, 4.9 million Canadians aged 15 and over experienced a need for mental health care; 1.6 million felt their needs were partially met or unmet~\cite{statscan}.
In 2017, over a third of Ontario students, grades 7 to 12, reported having wanted to talk to someone about their mental health concerns but did not know who to turn to~\cite{boak2017mental}.
These numbers highlight a concerning but all too familiar notion: although highly prevalent, mental health concerns often go unheard.
Nonetheless, mental disorders can shorten life expectancy by 7-24 years~\cite{chesney2014risks}.

In particular, depression is a major cause of morbidity worldwide. Although prevalence varies widely, in most countries, the number of persons that would suffer from depression in their lifetime falls between 8 and 12\%~\cite{epidemiology}.
Access to proper diagnosis and care is overall lacking because of a variety of reasons, from the stigma surrounding seeking treatment~\cite{stigma} to a high rate of misdiagnosis~\cite{misdiagnosis}. 
These obstacles could be mitigated in some way among social media users by analyzing their output on these platforms to assess their risk of depression or other mental health afflictions. 
The analysis of user-generated content could give valuable insights into the users mental health, identify risks, and help provide them with better support~\cite{ayers2013seasonality,choudhury2013predicting}.
To promote such analyses that could lead to the development of tools supporting mental health practitioners and forum moderators, the research community has put forward shared tasks like CLPsych~\cite{clpsych.org} and the CLEF eRisk pilot task~\cite{early.irlab.org,losadaetal2018}. Participants must identify users at risk of mental health issues, such as eminent risk of depression, post traumatic stress disorder, or anorexia. These tasks provide participants with annotated data and a framework for testing the performance of their approaches.

In this paper, we present a neural approach to identify social media users at risk of depression from their writings in a subreddit forum, in the context of the eRisk 2018 pilot task. 
From a technical standpoint, the principal interest of this investigation is the use of different aggregation methods for predictions on groups of documents.
Using the power of \glspl{rnn} for the sequential treatment of documents, we explore several manners in which to combine predictions on documents to make a prediction on its author.

\section{Dataset}

The dataset from the eRisk2018 shared task~\cite{losadaetal2018} consists of the written production of {\texttt{reddit}}~\cite{reddit} English-speaking users.

The dataset was built using the writings of 887 users, and was provided in whole at the beginning of the task. Users in the RISK class have admitted to having been diagnosed with depression; NO\_RISK users have not. 
It should be noted that the users' writings, or posts, may originate from different separate discussions on the website. The individual writings, however, are not labelled. 
Only the user as a whole is labelled as RISK or NO\_RISK.
The two classes of users are highly imbalanced in the training set with the positive class only counting 135 users to 752 in the negative class.
Table~\ref{tab:dataset} presents some statistics on the task dataset.

We use this dataset but consider a simple classification task, as opposed to the early-risk detection that was the object of the shared task.

\begin{table}
\centering
\setlength{\tabcolsep}{0.5em} 
{\renewcommand{\arraystretch}{1.5}
\begin{tabular}{r|rr|rr} 
& \multicolumn{2}{c}{training} & \multicolumn{2}{c}{test}\\
						& \bf risk 	& \bf control & \bf risk & \bf control	 \\ \hline
\# users	& 135 & 752 & 79 & 741\\
\# writings & 49,557 & 481,837 & 40,665 & 504,523\\
submissions / subject & 367.1 & 640.7 & 514.7 & 680.9\\
words / submission & 27.4 & 21.8 & 27.6 & 23.7
\end{tabular}
}
\vspace{.4cm}
\caption{Statistics on the eRisk 2018 task dataset}
\label{tab:dataset}
\end{table}

\section{Models}

We represent users as sets of writings rather than sequences of writings.
This is partly due to the intuition that the order of writings would not be significant in the context of forums, generally speaking.
It is also due to the fact that treating writings sequentially would be cumbersome, especially if we consider training on all ten chunks.
However, we do consider writings as sequences of words, as this is the main strength of \glspl{rnn}.
We therefore write a user $u$ as the set of his $m$ writings, $u=\{\mathbf{x}^{(1)}, \ldots, \mathbf{x}^{(m)}\}$.
A given writing $\mathbf{x}^{(j)}$, is then a sequence of words, $\mathbf{x}^{(j)} = x^{(j)}_1, \ldots, x^{(j)}_\tau$, with $\tau$ being the index of the last word.
Thus, $x^{(j)}_t$ is the $t$-th word of the $j$-th post for a given user.

\subsection{Aggregating predictions on writings}
\subsubsection{Late Inter-Document Averaging}
We set out to put together an approach that aggregates predictions made individually and sequentially on the writings of a user.
That is, we read the different writings of a user in parallel and take the average prediction on them. 
This is our first model, \gls{lida}.
Using the \gls{rnn} architecture of our choice, we read each word of a post and update its \emph{hidden state}, 
\begin{equation}\label{eq:h}
    h_t^{(j)} = f(x^{(j)}_t, h^{(j)}_{t-1}; \theta_{post}).
\end{equation}
$f$ is the \emph{transition function} of the chosen \gls{rnn} architecture, $\theta_{post}$ is the set of parameters of our particular \gls{rnn} model and the initial state is set to zero,
\begin{equation*}
    h_0 = \mathbf{0}.
\end{equation*}
In practice, however, we take but a sample of users' writings and trim overlong writings (see Sec.\ref{sec:meth}).
\Gls{lida} averages over the final state of the \gls{rnn}, $h^{(j)}_\tau$, across writings,
\begin{equation}
    a = \frac{1}{m}\sum_{j=1}^m h^{(j)}_\tau
\end{equation}
This average is then projected into a binary prediction for the user,
\begin{equation}\label{eq:pred}
    p = \sigma(\mathbf{u}^\top\left[\begin{array}{c}
         a \\
         1 
    \end{array}\right]),
\end{equation}
using $\sigma$, the standard logistic sigmoid function, to normalize the output and a vector of parameters, $\mathbf{u}$.
By averaging over all writings, rather than taking the sum, we ensure that the number of writings does not influence the decision.
However, we suspect that regularizing on the hidden state alone will not suffice, as the problem remains essentially the same: gradient correction information will have to travel the entire length of the writings regardless of the corrections made as a results of other writings. 

\subsubsection{Continual Inter-Document Averaging}
Our second model, \gls{cida}, therefore aggregates the hidden state across writings at \emph{every} time step, as opposed to only the final one.
A first \gls{rnn}, represented by its hidden state $h_t$, reads the writings as in Eq. \ref{eq:h}. 
The resulting hidden states are averaged across writings and then fed as the input to a second \gls{rnn}, represented by $g_t$,
\begin{equation}\label{eq:avg}
    a_t = \frac{1}{m}\sum_{j=1}^m h^{(j)}_t,
\end{equation}
\begin{equation}\label{eq:g}
    g_t = f(a_t, g_{t-1}; \theta_{user}).
\end{equation}
$g_\tau$ is used to make a prediction similarly to Eq.\ref{eq:pred}.

\subsection{Inter-document attention}

It stands to reason that averaging over the ongoing summary of each document would help in classifying a group of documents.
Nonetheless, one would suspect that some documents would be more interesting than others to our task.
Even if all documents were equally interesting, their interesting parts might not align well.
Because we are reading them in parallel, we should try and prioritize the documents that are interesting at the current time step.

\gls{cida} does not offer this possibility, as no weighting of terms is put in place in Eq.\ref{eq:avg}.
Consequently, we turn to the \emph{attention mechanism}~\cite{bahdanau2014neural} to provide this information.
While several manners of both applying and computing the attention mechanism exist \cite{luong2015effective,cheng2016long,showand}, we compute the variant known as \emph{general} attention \cite{luong2015effective}, which is both learned and content-dependent.
In applying it, we introduce \gls{ida}, which will provide a weighted average to our previous model.

The computation of $h^{(j)}_t$, the post-level hidden state, remains the same, i.e. Eq.\ref{eq:h}.
However, these values are compared against the previous user-level hidden state to compute the relevant \emph{energy} between them, $\hat{\alpha}_{jt}$
\begin{equation}\label{eq:alpha_at}
    \tilde{\alpha}^{(j)}_{t} = g_{t-1}\mathbf{W}_{att} h^{(j)}_t,
\end{equation}
where $\mathbf{W}_{att}$ is a matrix of parameters that learns the compatibility between the hidden states of the two \glspl{rnn}.
The resulting energy scalars, $\hat{\alpha}^{(j)}{t}$ are mapped to probabilities by way of softmax normalization,
\begin{equation}
    \alpha^{(j)}_t = \frac{e^{\tilde{\alpha}_t^{(j)}}}{\sum_{k=1}^me^{\tilde{\alpha}_t^{(k)}}}.
\end{equation}
This probability is then used to weight the appropriate $h_t$,
\begin{equation}
    a_t = \sum_{j=1}^m \alpha^{(j)}_t h^{(j)}_t.
\end{equation}
$g_t$ is given by Eq.\ref{eq:g}.
Through the use of this probability weighting, we can understand $a_t$ as an expected document summary at position $t$ when grouping documents together.
As in the previous model, a prediction on the user is made from $g_\tau$.

\subsection{Intra-document Attention}

We extend our use of the attention mechanism in the aggregation to the parsing of individual documents. 
Similarly to our weighting of documents in aggregation dependent on the current aggregation state, we compare the current input to past inputs to evince a \emph{context} for it.
This is known in the literature as \emph{self-attention}~\cite{cheng2016long}.
We therefore modify the computation of $h_t$ from Eq.\ref{eq:h} by adding a context vector, $c^{(j)}_t$, corresponding to the ongoing context in document $j$ at time $t$:
\begin{equation}\label{eq:h'}
    h_t^{(j)} = f(x^{(j)}_t, c^{(j)}_t, h^{(j)}_{t-1}; \theta_{post}).
\end{equation}
This context vector is computed by comparing past inputs to the present document-level hidden state,
\begin{equation}
    \tilde{\alpha}^{(j)}_{t, t'} = h^{(j)}_t\mathbf{W}_{intra}x^{(j)}_{t'} ,
\end{equation}
This weighting is normalized by softmax and used in adding the previous inputs together.
We refer to this model as \gls{iida}.

This last attention mechanism arises from practical difficulties in learning long-range dependencies by \glspl{rnn}.
While \glspl{rnn} are theoretically capable of summarizing sequences of arbitrary complexity in their hidden state, numerical considerations make learning this process through gradient descent difficult when the sequences are long or the state is too small~\cite{bengio1994learning}.
This can be addressed in different manners, such as gating mechanisms~\cite{lstm,gru} and the introduction of multiplicative interactions~\cite{mrnn}.
Self-Attention is one such mechanism where the context vector acts as a reminder of past inputs in the form of a learned expected context.
It can be combined to other mechanisms with minimal parameter load.

\section{Related Work}
Choudhury et al.~\cite{choudhury2013predicting} used a more classical approach to classify Twitter users as being at risk of depression or not.
They first manually crafted features that describe users' online behavior and characterize their speech.
The measures were computed daily, so a user is represented as the time series of the features.
Then, the training and predictions were done by a \gls{svm} with PCA for dimensionality reduction.

More similarly to our approach, Ive et al.~\cite{ive2018hierarchical} used Hierarchical Attention Networks~\cite{yang2016hierarchical} to represent user-generated documents.
Sentence representations are learned using a \gls{rnn} with an attention mechanism and are then used to learn the document's representation using the same network architecture.
The computation of the attention weights they use is different from ours as it is non-parametric.
Their equivalent of equation~\ref{eq:alpha_at} would be
\begin{equation}
    \tilde{\alpha}^{(j)}_t =
    h^{(j)\top}_{t} g_t^{(j)}
\end{equation}
This means that the \glspl{rnn} learn the attention weights along with the representation of the sequences themselves.
This attention function has been introduced in~\cite{luong2015effective} under that name of \textit{dot}.

The \textit{location-based function}~\cite{luong2015effective} is a simpler version of the \textit{general attention} that we used, that only takes into account the target hidden state. It is stated as such :
\begin{equation}
    \tilde{\alpha}^{(j)}_t =
    W_{att} g_t^{(j)}
\end{equation}

The \textit{additive} function introduced in~\cite{bahdanau2014neural}, has been improved in~\cite{luong2015effective}.
Luong et al. use a concatenation layer to combine the information of the hidden state and the context vector.
\begin{equation}
    \tilde{\alpha}^{(j)}_t =
    \text{tanh}(W_{att} g_t^{(j)} + W_{att} h_t^{(j)})
\end{equation}

\textit{Content-based addressing} was developed as part of Neural Turing Machines~\cite{graves2014neural}, where the attention is focused on inputs that are similar to the values in memory.
\begin{equation}
    \tilde{\alpha}^{(j)}_t =
    \text{cosine}[g_t^{(j)}, h_t^{(j)}]
\end{equation}

\section{Methodology}

\label{sec:meth}

\subsection{Preprocessing}

As previously mentioned, documents are broken into words.
The representation of these words is learned from the entirety of the training documents, all chunks included, using the skip-gram algorithm~\cite{mikolov2013distributed}.
All words were turned to lowercase.
Only the 40k most frequent words were kept.
The embedded representation learned is of size 40, using a window of size five.
The embeddings are are shared by all models.

Documents are trimmed at the end at a length of 66 words, which is longer than 90\% of the posts in the dataset. 
The number of documents varies greatly across user classes.
We take small random samples without replacement of 30 documents per user at every iteration (epoch).
We contend that sampling the user at every iteration allows us to train for longer as it is harder for the models to overfit when the components that make up each instance keep changing.

\subsection{Model configurations}

We use the \gls{mlstm}~\cite{mlstm} architecture as the post-level and user-level \gls{rnn}, where applicable.
The flexibility of the transition function in \gls{mlstm} has shown to be capable of arriving at highly abstract features on its own and can achieve competitive results in sentiment analysis~\cite{radford2017learning}.
Due to the limited number of examples, smaller models are required to avoid overfitting.
We therefore set the embedded representation at 20 and the size of the hidden state of both \glspl{rnn} to 80. 
Parameter counts are shown in Table~\ref{tab:results}.

\subsection{Training}

For our experiments, we reshuffle the original eRisk 2018 dataset, as the training and test sets do not have the same proportions among labels. 
To provide our models with more training examples, we divide the dataset 9:1, stratifying across labels.
We use 10\% of the training set as validation.

We train the models using the Adam optimizer~\cite{kingma2014adam}, making use of 10\% of the training data for validation.
Having posited random intra-user sampling as a means of training longer, we set the training time to 30 epochs, taking the best model on validation over all epochs. 
As noted, the two classes are highly imbalanced.
We use inverse class weighting to counteract this. 

\subsection{Evaluation}

The nature of the task, which is to prioritize finding positive users, and the class imbalance in the dataset, we use the f1-score as a first metric in validation and in the final testing phase.
The f1-score is useful to assess the quality of classification between unbalanced two unbalanced classes, one of which is designated as the positive class.
It is defined as the harmonic mean between \textit{precision} (out of all the positive examples, how many are correctly classified as positive) and \textit{recall} (out of all examples classified as positive, how many were indeed in the positive class).
Using True Positive (TP) as the number of positive examples correctly classified, False Positive (FP) the number of examples in the positive class incorrectly classifed, and True Negative (TN) and False Positive (FP) for the negative class, we have the following equations.\\
\begin{equation}
precision = \frac{TP}{TP + FP}
\end{equation}
\begin{equation}
recall = \frac{TP}{TP + FN}
\end{equation}
\begin{equation}
\text{\textit{f1-score}} = 2 \times \frac{precision \times recall}{precision + recall}
\end{equation}

We evaluate our models on the best result on a validation set of 10\% of the training data.
These best results are selected over 30 epochs.

\section{Results}

\begin{table}
\centering
\setlength{\tabcolsep}{0.5em} 
{\renewcommand{\arraystretch}{1.5}
\begin{tabular}{r|r|r|r|r} 
model & p.c. & precision & recall & f1-score \\ 
\hline
\gls{lida}	& 31k  & 39.7 & 51.2 & 45.6 \\
\gls{cida} & 95k  & 41.7 & 69.8 & 52.2 \\
\gls{ida}  & 101k  & 45.6 & 73.2 & 56.2 \\
\gls{iida}  & 175k  & 47.4 & 72.8 & 57.4 \\
\end{tabular}}
\vspace{.4cm}
\caption{Parameter counts, precision, recall and f1-score (\%) on the adapted test set for the eRisk 2018 corpus}
\label{tab:results}
\end{table}

Our preliminary results in validation are in accordance with our hypotheses.
That is, continual aggregation surpasses late aggregation but falls short of the more sophisticated attention model.
Moreover, the noticeable difference in performance has little to no cost in terms of parameter count.


\section{Conclusion}
In this paper, we have put forward three \gls{rnn}-based models that aggregate documents to make a prediction on their author.
We applied this model to the eRisk 2018 dataset, which associates a user, as a sequence of online forum posts, to a binary label that identifies them as being at risk for depression or not.

With the goal of using \glspl{rnn} to read the individual documents, we tested four methods of combining the resulting predictions, \gls{lida}, \gls{cida}, \gls{ida} and \gls{iida}.
We also introduced the inter-document attention mechanism.
Our preliminary results show promise and confirm the parameter efficiency of the attention mechanism.

Future work could involve the use of dot-product alone, which, despite adding no parameters, has been found to be more effective for global attention~\cite{luong2015effective}.
An investigation into using late attention aggregation for all hidden states produced across all documents is also necessary.

\bibliography{ref}
\bibliographystyle{splncs04}
\end{document}